\documentclass[letterpaper,twocolumn,10pt]{article}
\usepackage{usenix}

\usepackage[utf8]{inputenc} 
\usepackage[T1]{fontenc}    
\usepackage{hyperref}       
\usepackage{url}            
\usepackage{authblk}        
\usepackage{color}          
\usepackage[numbers]{natbib}

\usepackage{multirow}       
\usepackage{booktabs}       
\usepackage{amsmath}        
\usepackage{amsthm}         
\usepackage{wrapfig}        
\usepackage{subcaption}     
\usepackage{enumitem}       

\usepackage{amsfonts}       
\usepackage{nicefrac}       
\usepackage{microtype}      
\usepackage{stmaryrd}       
\usepackage{amssymb}        

\usepackage[n,advantage, operators, sets, adversary, landau, probability, notions, logic, ff, mm, primitives, events, complexity, oracles, asymptotics, keys]{cryptocode}

\setlist{noitemsep}         

\newtheorem{lemma}{Lemma}

\DeclareSymbolFont{operators}   {OT1}{cmr} {m}{n}
\DeclareSymbolFont{letters}     {OML}{cmm} {m}{it}
\DeclareSymbolFont{symbols}     {OMS}{cmsy}{m}{n}
\DeclareSymbolFont{largesymbols}{OMX}{cmex}{m}{n}
\SetSymbolFont{operators}{bold}{OT1}{cmr} {bx}{n}
\SetSymbolFont{letters}  {bold}{OML}{cmm} {b}{it}
\SetSymbolFont{symbols}  {bold}{OMS}{cmsy}{b}{n}
\DeclareSymbolFontAlphabet{\mathrm}    {operators}
\DeclareSymbolFontAlphabet{\mathnormal}{letters}
\DeclareSymbolFontAlphabet{\mathcal}   {symbols}
\DeclareMathAlphabet      {\mathbf}{OT1}{cmr}{bx}{n}
\DeclareMathAlphabet      {\mathsf}{OT1}{cmss}{m}{n}
\DeclareMathAlphabet      {\mathit}{OT1}{cmr}{m}{it}
\DeclareMathAlphabet      {\mathtt}{OT1}{cmtt}{m}{n}

\begin{document}

\date{}

\title{\Large \bf 

Practical and Light-weight Secure Aggregation for
Federated Submodel Learning
}


\author[1]{\textbf{Jamie Cui}}
\author[2*]{\textbf{Cen Chen}}
\author[2]{\textbf{Tiandi Ye}}
\author[1]{\textbf{Li Wang}}
\affil[1]{Ant Group}
\affil[2]{East China Normal University}
\affil[*]{Corresponding author, email: cenchen@dase.ecnu.edu.cn}

\maketitle

\begin{abstract}
Recently, Niu, et. al. \cite{niu2020billion} introduced a new variant of Federated Learning (FL), called \emph{Federated Submodel Learning} (FSL). Different from traditional FL, each client locally trains the submodel (e.g., retrieved from the servers) based on its private data and uploads a submodel at its choice to the servers. Then all clients aggregate all their submodels and finish the iteration. Inevitably, FSL introduces two privacy-preserving computation tasks, i.e., \emph{Private Submodel Retrieval} (PSR) and \emph{Secure Submodel Aggregation} (SSA). Existing work fails to provide a loss-less scheme, or has impractical efficiency. In this work, we leverage \emph{Distributed Point Function} (DPF) and cuckoo hashing to construct a practical and light-weight secure FSL scheme in the two-server setting. More specifically, we propose two basic protocols with few optimisation techniques, which ensures our protocol practicality on specific real-world FSL tasks. Our experiments show that our proposed protocols can finish in less than 1 minute when weight sizes $\leq 2^{15}$, we also demonstrate protocol efficiency by comparing with existing work \cite{niu2020billion} and by handling a real-world FSL task.


\end{abstract}

\section{Introduction}
\label{sec:introduction}

\emph{Federated Learning} (FL) is a distributed machine learning paradigm, where participants jointly train a global model based on their private data without sharing them~\cite{mcmahan2017communication}. FL ensures data privacy to a certain extend by restricting the communication on only exchanging the model parameters.
Specifically, in each communication round, the selected clients will retrieve the current model from the global server, locally train the models based on their respective private data, and upload the updated model parameters for global aggregation.
However, directly sharing model parameters may leak unintended information about participants’ training data \cite{melis2019exploiting}. To enhance privacy, protocols based on \emph{secure aggregation} are designed for federated learning \cite{bonawitz2017practical,kadhe2020fastsecagg,bell2020secure,fereidooni2021safelearn}. The basic idea is to leverage \emph{Multi-Party Computation} (MPC) to calculate the sum of the updates in a secure manner, so that the global server only recovers the aggregated model updates.

%

In the literature, most of the existing works focus on the secure aggregation of the whole model regarding the individual weight updates \cite{bonawitz2017practical,kadhe2020fastsecagg,bell2020secure,fereidooni2021safelearn}.
Recently, Niu, et. al.\cite{niu2020billion} introduces the concept of \emph{Federated Submodel Learning} (FSL), where each client only retrieves and trains a submodel from the servers.
FSL is industry-motivated, with the aim of providing more accurate, customized and scalable model training than traditional FL. 
%
%
However, there are concerns that each client's selection of submodel may contain sensitive data.
For example, in natural language processing (NLP), a client's selection points out its desired word embedding vectors.
That is to say, a corrupted server can retrieve the client's words by looking at the selection positions in its vocabulary.

In this paper, we try to answer the following questions:
\begin{enumerate}
    \item \emph{Can we allow each client to privately retrieve a submodel from the aggregation servers?}
    \item \emph{Can we allow the servers to securely aggregate clients' submodel weight updates?}
\end{enumerate}

Trivially, we can allow each client first retrieves the entire model (instead of a submodel) from the servers, and runs the de facto secure aggregation protocol in FL (e.g. the protocol by Keith, et. al. \cite{bonawitz2017practical}) to updates the entire model.
To make those questions non-trivial, we want a protocol that achieves better asymptotic communication complexity than the best known secure aggregation protocols in the literature of FL.

Observe that the first question has close relationship to a well-known problem, \emph{Private Information Retrieval} (PIR).
Intuitively, PIR protocols allow a client to privately retrieve an entry from the server's database while ensuring that the server learns nothing about which entry is retrieved.
The trivial answer of PIR is to let the server send the entire database to the client, and the client locally selects its desired data.
Hence we say a PIR protocol is non-trivial if it is communication-wise better than the trivial solution.
In the literature, many efficient PIR protocols have been proposed, either in single-server setting \cite{angel2018pir, ali2019communication}, or in multi-server setting \cite{corrigan2020private, canetti2017towards, gilboa2014distributed}.
As for the second question, similarly, we say a secure submodel aggregation protocol is non-trivial if it is concretely better than the full model secure aggregation protocol in terms of communication.
Existing solutions \cite{niu2020billion, beguier2020safer, jia2020x, kim2020information} leverages different tools, i.e. \emph{Differential Privacy} (DP) \cite{dwork2014algorithmic, mironov2009computational}, \emph{Multi-Party Computation} (MPC) \cite{damgaard2012multiparty} or PIR \cite{angel2018pir,ali2019communication,corrigan2020private, canetti2017towards, gilboa2014distributed}.
However, some work~\cite{kim2020information, jia2020x} consider only the non-adaptive case, where each client selects a fixed submodel prior to the training.
Other work \cite{niu2020billion, beguier2020safer} introduce calibrated training noises to preserve privacy.
There is lack of study of generic techniques for secure submodel aggregation.
%

%
In this paper, we work in a setting where each client only interacts with \emph{two non-colluding servers}.
%
We say an aggregation protocol is secure in FSL if it preserves the privacy of each client's submodel and the submodel updates in each communication round.
%
We consider the \emph{threat model} with at least one of the servers is honest (the other may deviate arbitrarily from the protocol and may collude with an unbounded number of malicious clients).
%
For example, one could choose servers from two different cloud service providers (CSPs) (such as Google and Amazon), or one from CSPs and one from trusted authorities.
In the meantime, with such a two-server model, we are able to answer the two questions using only light-weighted cryptographic tools, without heavy public-key operations \cite{cheon2017homomorphic, fan2012somewhat}, and generic secure computation techniques \cite{damgaard2012multiparty, keller2018overdrive}.
By leveraging only light-weighted tools, our protocol is concretely efficient, and practical.

As a result, we propose two basic protocols for both private retrieval and secure aggregation.
Our basic protocols only leverage \emph{Pseudo-Random Generator} (PRG) and require secure P2P channels between clients and servers, and between both servers.
In fact, each client in our protocol sends only one message to both servers during either private retrieval or secure aggregation.
Roughly speaking, if we denote $\lambda$ as the security parameter, $k$ as the average submodel size for all clients, and $l$ as the bit-length of weight value, each client has communication complexity of $O(k(\lambda + 1))$ and $O(k(\lambda + l))$, respectively.
More concretely, for MNIST dataset training (4 CNN layers and 2 fully-connected layers, 1,663,370 weights) and 1\% top weight selection, our method has approximately $12\%$ compression effect on communication at client side.
Decreasing the communication from 203MB to 24MB.
As for computation complexity, each client invokes $O(k)$ and $O(kl)$ AES encryptions in the counter mode separately.
As for the MNIST example, each client can finish its computation within 1~minute.
Apart from the basic protocols, we also propose several optimisations aiming at different scenarios.
(1) Integrating \emph{Private Set Union} (PSU), this method works in the scenario where the union set of all client's submodels is still much smaller than the entire model.
(2) We also introduce the concept and construction of Updatable \emph{Distributed Point Function} (DPF), which allows each client to update its weight update at a cheaper communication cost when each client's submodel is fixed during one training task.
(3) Submodel with mega-element grouping, this optimisation allows a client to group elements into mega-elements.
%


\medskip
\noindent
\textbf{Limitations}.
Our proposed basic protocol (without any optimisations) is only non-trivial when $k/m\lessapprox 10\%$.

\medskip
\noindent
\textbf{Contributions}.
We summarize our contributions as follows,
\begin{enumerate}
    \item We propose the first \emph{practical} and \emph{lossless} secure FSL aggregation framework against a single malicious server and arbitrary number of malicious clients.
    
    \item Our proposed method is a novel combination of \emph{Cuckoo Hashing} and \emph{Distributed Point Function} (DPF). We also proposed a slightly-modified variant of DPF called Updatable DPF (U-DPF).
    
    \item Our proposed \emph{basic} protocols are generic secure submodel aggregation techniques, which can be applied to either adaptive or non-adaptive cases. Also, our protocols are practical conditioned on small $k/m$.
    
    \item 
    Several \emph{optimisations} are further proposed based on the basic version of our protocol, and demonstrate the significant efficiency improvements through empirical experiments.
\end{enumerate}

\smallskip
\noindent
\textbf{Organization}.
In section \ref{sec:problem-formulation}, we introduce the problem of PSR and SSA, and then formalize the functionality and security definitions.
Section \ref{sec:preliminaries} introduces the cryptographic tools we use in our protocol construction, including DPF and batch codes.
Next, in section \ref{sec:basic-protocols}, we introduce our basic protocols of PSR and SSA, and shows the limitations of the basic protocols.
Then in section \ref{sec:optimisations} we proposed several optimization techniques.
Finally in section \ref{sec:experiments}, we run empirical experiments to demonstrate the efficiency of our protocols.

\section{Problem Formulation}
\label{sec:problem-formulation}

We assume there are \emph{two non-colluding servers}, $\mathcal{S}_0$ and $\mathcal{S}_1$, which jointly perform secure computation tasks.
Also, there are $n$ clients, where we denote the $i$-th client as $\mathcal{C}_i$.
In addition, for all $0\leq i<n$, we assume that there are secure P2P communication channels between: (1) $\mathcal{S}_0$ and $\mathcal{S}_1$; (2) $\mathcal{C}_i$ and $\mathcal{S}_0$; (3) $\mathcal{C}_i$ and $\mathcal{S}_1$.

\medskip
\noindent
\textbf{Notations}. 
In this paper, we use $\GG$ to denote a finite Abelian group and $\FF$ to denote a prime field.
We also use $\sample$ to denote the sampling of uniformly random elements, and denote the assignment of variables as $x\gets 4$.
Moreover, we use capital letter to denote a table (except for model weights and the submodel selections, which are denoted as $\textbf{w}$ and $\textbf{s}$ separately). We also use $T[i]$ to denote the $i$-th element in the table $T$
and use $|\cdot|$ to denote the size of an element.
We summarize other notations in Table \ref{tab:notation}.

\begin{table}[t!]
    \centering
    \begin{tabular}{@{}ll@{}}
    \toprule
    \textbf{Symbol} & \textbf{Descriptions}\\
    \midrule
    $m$ & number of the global weights\\
    $k\leq m$ & number of client's selected weights\\
    $n$ & number of clients\\
    $\lambda$ & computational security parameter\\
    $\kappa$ & statistical security parameter\\
    $\epsilon$ & cuckoo parameter: scale factor\\
    $\eta$ & cuckoo parameter: hash numbers\\
    $\sigma$ & cuckoo parameter: stash size\\
    $\GG$ & a finite Abelian group\\
    $\textbf{s}^{(i)}\in\ZZ_m^k$     &  client $i$'s selected submodel indexes\\
    $\textbf{w}\in\GG^m$     &  global model weights\\
    $\textbf{w}^{(i)}\in\GG^k$     &  client $i$'s desired model weights\\
    $\Delta\textbf{w}\in\GG^m$     &  global model weight updates\\
    $\Delta\textbf{w}^{(i)}\in\GG^k$     &  client $i$'s local trained model update\\
    \bottomrule
    \end{tabular}
    \caption{Notations}
    \label{tab:notation}
\end{table}

\subsection{Federated Submodel Learning}
\begin{figure}[t!]
    \centering
    \includegraphics[width=\linewidth]{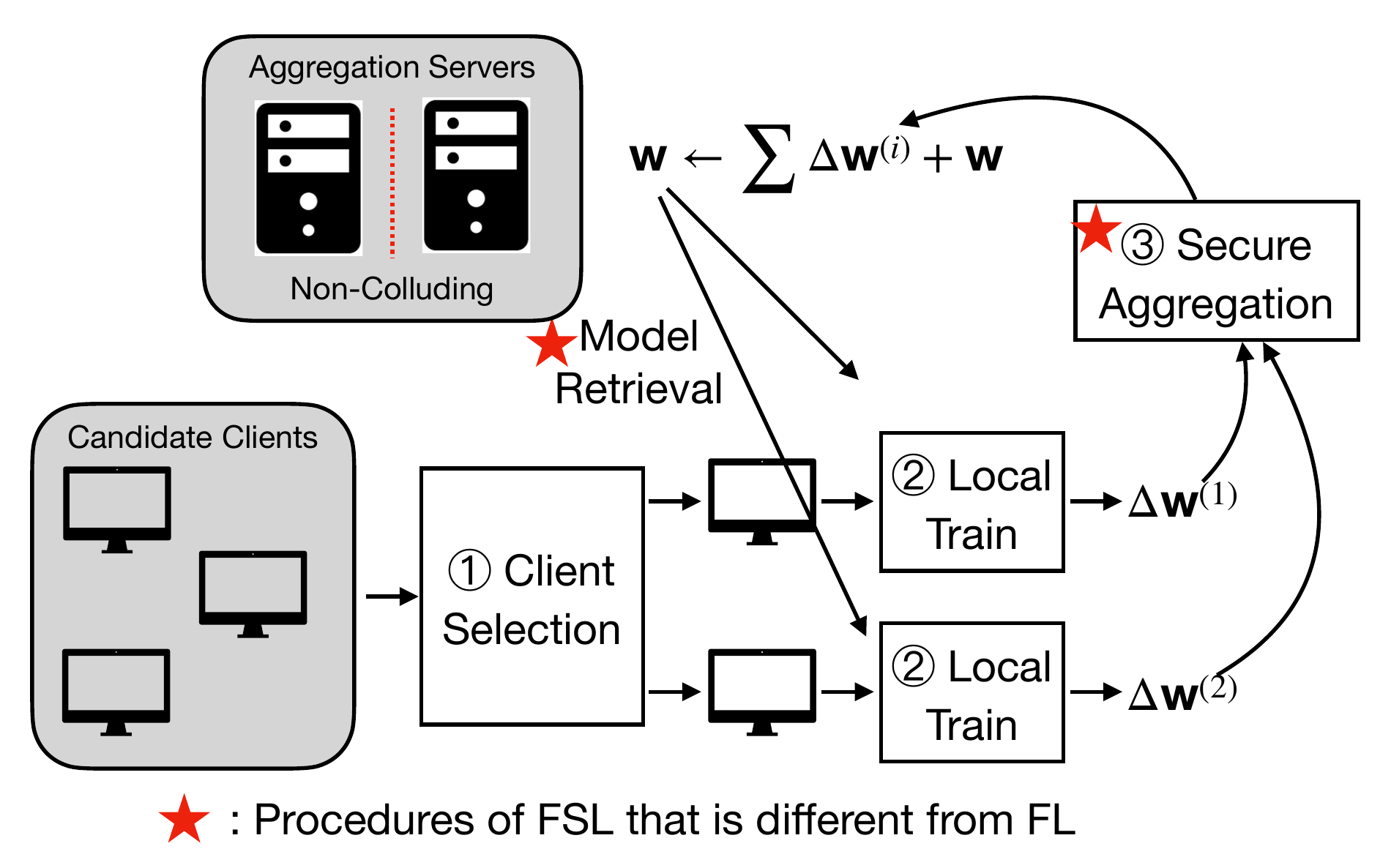}
    \caption{An overview of federated submodel learning system}
    \label{fig:fsl}
\end{figure}

Federated submodel learning is an efficient learning paradigm arising from the federated scenarios where individual local client's data is only pertinent to a small portion of the full model being trained, thus, submodels are sent for aggregation to further improve communication efficiency \cite{niu2020billion}.
Specifically, in each communication round, first, parties jointly selects $n$ clients that will participant in this training round.
To simply the notion, we refer the participant clients as clients.
Each client then retrieves a submodel of his choice from the servers.
Here, the submodel of each client may contain sensitive information.
For instance, in e-commerce recommendation, a client’s submodel mainly contains the embedding vectors for the goods IDs in its historical data, and also the parameters of the other network layers. 
Afterwards, each client locally trains the submodel over its private data and gets the submodel update.
All clients then jointly invoke a secure aggregation protocol and gets the final global model update $\Delta\textbf{w}$.
The secure aggregation protocol ensures than parties learn nothing except its own input and the global model update.
We show an overview of FSL system in Figure \ref{fig:fsl}.
In summary, a secure FSL framework ought to meet two essential requirements, i.e., (1) hides each client's submodel, including selections and values and (2) ensures the correctness of FSL functionality, which are closely related to the following tasks.
%

\medskip
\noindent
\textbf{Task 1: Private Submodel Retrieval (PSR).}
We defined this task between clients and servers.
More specifically, we assume the servers hold a size-$m$ vector $\textbf{w}\in\GG^m$, where $\GG$ is an Abelian group.
Each client independently holds $k$ non-repeat queries, where we denote the queries for client $i$ as $\textbf{s}^{(i)}$.
For each query $s\in\textbf{s}^{(i)}$, client $i$ want to learn the exact value of $\textbf{w}_s$, while hiding each selection $s$ from the servers.
Since we allow each client to learn the entire vector $\textbf{w}$, the PSR problem can be reduced to multi-query private information retrieval (PIR) problem.
Intuitively, a PSR protocol is non-trivial only if it is concretely more efficient than downloading the entire $\textbf{w}$.
In reality, FL clients usually have sufficient download bandwidth (for example, 100 MB) but limited upload bandwidth (for example, 10 MB).
%
Therefore, the PSR problem is less significant when considering downlink communication compared to uplink.
%

\medskip
\noindent
\textbf{Task 2: Secure Submodel Aggregation (SSA).}
Similar to PSR, we also define SSA between $n$ clients and two servers.
To begin with, consider all participants share the same zero list with size-$m$, i.e. $\Delta\textbf{w}\in\set{0}^m$.
Each client then holds two list representing the submodel indexes and the submodel weight updates.
More specifically, we assume client $i$ holds submodel indexes $\textbf{s}^{(i)}\in\ZZ_m^k$ and submodel weight updates $\Delta\textbf{w}^{(i)}\in\GG^k$.
That is to say, both servers in SSA should be instructed by the client to aggregate each value $\Delta w^{(i)}_j$ at the $s^{(i)}_j$-th position of $\Delta\textbf{w}$ in an oblivious way.
The obliviousness of SSA meets the security requirement in the sense that the servers should learn negligible information about client's submodel, that is, $\textbf{s}^{(i)}$ and $\textbf{w}^{(i)}$.
To make the SSA problem non-trivial, observe that we can let each client expands their submodel to size-$m$ with zeros, and invokes the full-blown secure aggregation protocol to perform submodel aggregation.
Therefore, we say a SSA protocol is non-trivial if it has better asymptotic communication complexity or is concretely communication-wise better than the trivial SSA.
%

\subsection{Security Definition}

Recall the secure FSL is a sub-variant of federated learning, the security of FL preserves if during one iteration, parties learn nothing except the model updates for this round.
In the literature, we could leverage \emph{Differential Privacy} (DP) to add additional protection to each rounds' model updates.
We believe the discussion of this is of independent interest.
In this work, we use simulation-based security definition.
This means that we need to specify a precious ideal functionality for each type of corruption.

\medskip
\noindent
\textbf{PSR Functionality.}
In FSL, each client independent invokes PSR with the two servers. 
Therefore, we only assume a client and two servers in PSR.
Note it's trivial to consider a malicious client since PSR aims to ensure the security of client's submodel selection.
\begin{itemize}
    
    \item \textit{Participants}: a client $\mathcal{C}$, two servers $\mathcal{S}_0, \mathcal{S}_1$.
    \item \textit{Functionality for honest parties (ideal world)}:
    \begin{enumerate}
        \item Receives $\textbf{w}\in\GG^m$ from servers.
        \item Receives $\textbf{s}^{(i)}\in\set{1,\ldots, m}^k$ from $\mathcal{C}$.
        \item Returns to $\mathcal{C}$ with $w_j$ for all $j\in \textbf{s}^{(i)}$.
    \end{enumerate}
    
    \item \textit{Leakage for malicious server}: Suppose both Server $\mathcal{S}_b$ and client set $\mathcal{C}_\mathsf{cor}\in\mathcal{C}$ are controlled by an efficient adversary. The view of the adversary can be simulated given leakage function $\mathcal{L}=(k)$.
\end{itemize}

\medskip
\noindent
\textbf{SSA Functionality.} 
Different from PSR, SSA assumes the existence of $n$ clients and two servers.
\begin{itemize}
    \item \textit{Participants}: $n$ clients $\mathcal{C}_{1\leq i \leq n}$, two servers $\mathcal{S}_0, \mathcal{S}_1$.
    \item \textit{Functionality for honest parties (ideal world)}:
    \begin{enumerate}
        \item Receives all $\textbf{w}^{(i)}\in\GG^k$ from clients.
        \item Calculates $\textbf{w}\gets\mathsf{Agg}(\textbf{w}^{(1)}, \ldots, \textbf{w}^{(n)})$, where $\mathsf{Agg}$ is the aggregation functionality.
        \item Returns to the servers with $\textbf{w}\in\GG^m$.
    \end{enumerate}
    
    \item \textit{Functionality for malicious clients}: Suppose we denote the set of corrupted clients by adversary as $\mathcal{C}_\mathsf{cor}$, the influence of the adversary on the output is captured by the following functionality.
    \begin{enumerate}
        \item For each $\mathcal{C}_i\in\mathcal{C} \setminus \mathcal{C}_\mathsf{cor}$ (honest clients), receives all $\textbf{w}^{(i)}\in\GG^k$ from $\mathcal{C}_i$.
        \item For each $\mathcal{C}_i\in \mathcal{C}_\mathsf{cor}$ (corrupted clients), receives  $\textbf{w}^{(i)}_\mathsf{cor}\in\GG^k$ and selective vote predicate $V^{(i)}_\mathsf{cor}:\GG^k\to\bin$ from the adversary. For $\textbf{x}\in\GG^k$, we require that if $V^{(i)}_\mathsf{cor}(\textbf{x})=1$, then $\textbf{x}=\textbf{w}^{(i)}$. Intuitively, if we have $V^{(i)}_\mathsf{cor}(\textbf{x})=1$, the vote of $\mathcal{C}_i$ will not count.
        \item Returns to the servers with $\textbf{w}\in\GG^m$.
    \end{enumerate}
    
    \item \textit{Leakage for malicious server}: Suppose both Server $\mathcal{S}_b$ and client set $\mathcal{C}_\mathsf{cor}\in\mathcal{C}$ are controlled by an efficient adversary. The view of the adversary can be simulated given leakage function $\mathcal{L}=(k)$.
\end{itemize}

\section{Preliminaries}
\label{sec:preliminaries}

This section introduces the cryptographic primitives we use in our proposed protocol, including the concept of distributed point function and batch codes.

\subsection{Distributed Point Function (DPF)}
Distributed Point Function (DPF) \cite{gilboa2014distributed, boyle2015function, boyle2016function} allows a dealer to secretly share a point function to two parties, where two parties can jointly evaluate an input based on the point function.
A point function is defined as $f_{\alpha, \beta}:\bin^n\to\GG$, where $\alpha\in\bin^n$, $\beta\in\GG$.
The point function $f_{\alpha, \beta}$ evaluates to $\beta$ on input $\alpha$, and $0$ on all other inputs.
Formally, a DPF protocol has two algorithms $(\mathsf{Gen}, \mathsf{Eval})$, given $\alpha\in\bin^n$ and $\beta\in\GG$:
\begin{description}
    \item[$\mathsf{Gen}(1^\lambda, \alpha, \beta):$] Given security parameter $\lambda\in\NN$, a point $\alpha\in\bin^n$ and a value $\beta\in\GG$, output two DPF keys $(k_0, k_1)$ which jointly imply the point function $f_{\alpha, \beta}$.
    \item[$\mathsf{Eval}(b, k_b, x):$] Given a DPF key $k_b$ (where $b\in\bin$) and a point $x\in\bin^n$, output the secret-shared value at the position $x$ on $f_{\alpha,\beta}$.
\end{description}
The correctness of DPF says that, for all $\lambda, \alpha, \beta$, if we have $(k_0, k_1)\gets\mathsf{DPF.Gen}(1^\lambda, \alpha, \beta)$, and $x\in\bin^n$,
\begin{align*}
    \mathsf{DPF.Eval}(k_0, x) + \mathsf{DPF.Eval}(k_1, x) 
    \begin{cases}
        \beta\in\GG,& \text{if } x= \alpha\\
        0\in\GG,              & \text{otherwise}.
    \end{cases}
\end{align*}
Roughly, the security of DPF says the adversary holding either $k_0$ or $k_1$ learns nothing about $\alpha$ or $\beta$.
The best known DPF \cite{boyle2016function} construction has key size of $n(\lambda+2)+\lambda+\ceil{\log\GG}$ bits.
Observe in their DPF construction, each DPF keys have (1) an \emph{public part} with $n(\lambda+2) +\ceil{\log\GG}$ bits which is identical to both keys, and (2) a \emph{private part} with $\lambda$ bits which differs between the two DPF keys.
We give the construction details of DPF in the appendix.

\medskip
\noindent
\textbf{Malicious-secure sketching.}
Recall that we would like our protocol to be secure against any number of malicious clients and a malicious server, we would like a DPF scheme satisfying such security definition.
Previous works \cite{boneh2019zero, boyle2016function} present techniques to make standard DPF scheme against malicious client, and a recent work \cite{boneh2020lightweight} introduces malicious-secure sketching and extractable DPF to let it further against a malicious server.
Also they pointed out that the public-parameter variant of the DPF in \cite{boyle2016function} is an extractable DPF which is secure against malicious client.
Since the servers only run a ``sketching'' protocol in their protocol to check the validity of the DPF outputs when considering malicious adversaries, in the following, we omit the sketching check by servers.
Finally, we follow the sketching protocol of \cite{boneh2020lightweight}.

\subsection{Batch Codes}

Previously, batch codes are popularly used in the literature of \emph{Private Set Intersection} (PSI)~\cite{kolesnikov2016efficient,Kolesnikov2019ScalablePS, rindal2021vole} and \emph{Private Information Retrieval} (PIR)~\cite{angel2018pir,Kolesnikov2019ScalablePS}.
Informally, a batch code encodes a string into multiple bins, such that the string can be retrieved by reading at most one character from each bin.
Now, if we let $T[i]$ denotes the $i$-th bin in the table $T$, the batch code is defined as $\mathcal{B}:\{x_1,...,x_n\}\to\{T[1],...,T[m]\}$, where we have $m<n$, and $\sum_{i=1}^m |T[i]|\geq n$.
Since existing batch codes introduce significant communication overhead to reduce computation, here we use \emph{Probabilistic Batch Code} (PBC) which fails with a marginally probability $p$.

One of the most popular instantiations of PBC is cuckoo hashing.
Cuckoo hashing tables store (key, value) pairs with worst-case constant lookup.
Concretely, cuckoo hashing uses $\eta$ hash functions, $h_1,\cdots, h_\eta$, to insert $n$ elements into $m\leq n$ bins.
Typically one could set $\eta=3$ for most use cases.
Note that cuckoo hashing ensures that each key resides in one of the $\eta$ locations determined by one of the hash function evaluations on the key.
Hash collisions are resolved using the cuckoo approach: if a collision occurs when placing an item in the hash table, the item residing in the location is evicted and then placed in the table using a different hash function, potentially evicting another item in the case of collision. 
This process continues until all evicted items are placed, if possible.
If the resulting inserting sequence fails at a certain number of iterations, the current element is inserted into a special bin called stash.
In section \ref{sec:basic-protocols}, we leverage cuckoo hashing with a stash to construct our basic protocols for the sake of completeness.
Since the invocation of stash incurs enormous computational cost, later we adopt the stash-less setting (where the stash size is zero) in all our experiments.
\section{Basic Protocols}
\label{sec:basic-protocols}

In this section, we present the technical details of our basic protocols for \emph{Private Submodel Retrieval} (PSR) and \emph{Secure Submodel Aggregation} (SSA).
Our basic protocols leverage symmetric operations only and hence is light-weight and efficient.
Later in section \ref{sec:optimisations}, we introduce several optimisation techniques to make our solutions practical in real-world.

Recall that in each round, PSR assumes that both servers ($\mathcal{S}_0$ and $\mathcal{S}_1$) hold the previously aggregated model $\textbf{w}\in\GG^m$.
For $1\leq i\leq n$, client $\mathcal{C}_i$ holds a index list $\textbf{s}^{(i)}$ indicating the submodel that it want to retrieve.
Inspired by existing multi-query PIR protocols \cite{angel2018pir, ishai2004batch}, we use cuckoo hashing to break down the requirement of multi-query PIR to multiple invocations of single-query PIR protocols.
Then we leverage the existing two-server PIR protocol based on DPF \cite{gilboa2014distributed} to finish our basic PSR protocol.
In particular, we first let each $\mathcal{C}_i$ inserts its index set $\textbf{s}^{(i)}$ into a table using cuckoo hashing.
We denote the cuckoo table as $T_\mathsf{cuckoo}$.
Cuckoo hashing ensures that each bin in $T_\mathsf{cuckoo}$ is either empty or has exactly one element.
Similarly, all participants insert the full index set $\set{1,...,m}$ into a table using simple hashing, which is denoted as $T_\mathsf{simple}$.
Different from $T_\mathsf{cuckoo}$, bins in the simple table are allowed to contain multiple elements.
Moreover, we choose the same parameters for both simple hashing and cuckoo hashing (i.e. scale-factor, hash functions), which guarantees that for the $j$-th bin of both tables, the element $T_\mathsf{cuckoo}[j]$ belongs to the list $T_\mathsf{simple}[j]$.
In that way, we reduced the problem of how to retrieve $\textbf{s}^{(i)}$ from $\textbf{w}$, to how to retrieve $T_\mathsf{cuckoo}[j]$ from the list $\set{w_{T_\mathsf{simple}[j][0]},\ldots, w_{T_\mathsf{simple}[j][\Theta-1]}}$ for each $j$-th bin, here we use $\Theta$ to denote the maximum bin size of $T_\mathsf{simple}$.
To simplify the notation, for the $j$-th bin, we denote the index of element in $T_\mathsf{simple}$ that the client wants to retrieve as $\mathsf{pos}_j$, and the maximum bin size for $T_\mathsf{simple}$ as $\Theta$.
To retrieve the $\mathsf{pos}_j$-th element, we use the DPF-based two-server PIR protocol proposed by \cite{gilboa2014distributed}.
We let the client first generates two DPF keys $(k_0, k_1)$ which jointly implies the point function $f_{\mathsf{pos}_j, 1}$.
Then let the client send $k_b$ to $\mathcal{S}_b$ for $b\in\bin$.
Once $\mathcal{S}_{b}$ receives the DPF keys for the $j$-th bin, $\mathcal{S}_{b}$ evaluates on all possible inputs $x\in\ZZ_\Theta$ of the DPF function, and gets his secret-shared result $\set{[f_{\mathsf{pos}_j, 1}(x)]_b}_{x}$.
Recall the definition of point function, there is only one value of $x$ that is evaluated to $1$, all other values are evaluated to $0$.
Now let each $\mathcal{S}_{b}$ calculate $[w^\prime]_b\gets\sum_{x} w_x[f_{\mathsf{pos}_j, 1}(x)]_b$, and finally sends $[w^\prime]_b$ back to the client.
The correctness of our PSR protocol can be verified by the following,
\begin{align*}
    w_h=[w^\prime]_0 + [w^\prime]_1 \in\GG= \sum_{x}w_x f_{\mathsf{pos}_j,1}(x) \in \GG.
\end{align*}

\begin{figure}[t!]
    \centering
    \includegraphics[width=0.8\linewidth]{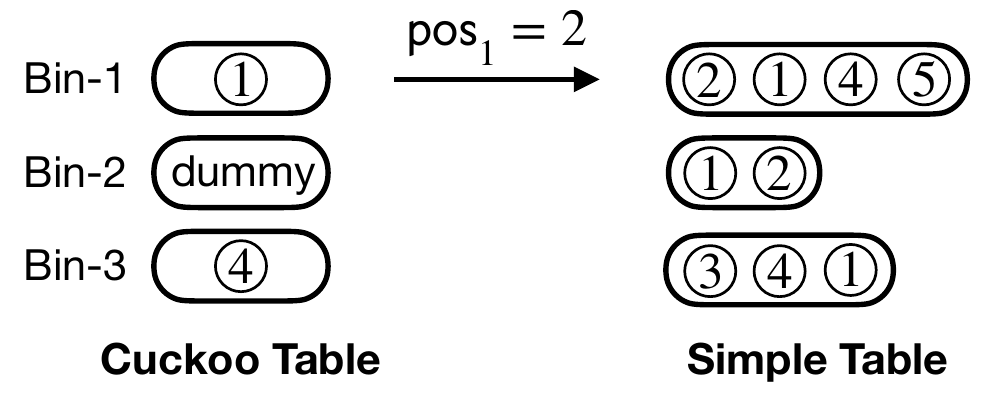}
    \caption{An example: cuckoo hashing and simple hashing work together, where we insert $\set{1,4}$ into cuckoo table and $\set{1,\ldots, 5}$ into simple table with 3 hash functions. Note that element ``2'' is only inserted into two bins in the simple table, this is due to the fact that it has at most one collision hash values (e.g. $h_2(2)$  may collides with $h_3(2)$).}
    \label{fig:cuckoo}
\end{figure}

As for secure submodel aggregation, recall that we assume each $\mathcal{C}_i$ holds two lists $(\Delta\textbf{w}^{(i)}\in\GG^k, \textbf{s}^{(i)}\in\ZZ_m^k)$.
Similarly, we could reuse $T_\mathsf{cuckoo}, T_\mathsf{simple}$ in PSR protocol since the submodel of each client remains the same.
For instance, in the $j$-th bin the client wants to select the $\mathsf{pos}_j$-th element (we denote this element as $u$) at the simple table.
Instead of constructing DPF keys implying $f_{\mathsf{pos}_j, 1}$, we let client generates DPF keys for the mapping $\mathsf{pos}_j\to \Delta w^{(i)}_u$.
Due to the property of the cuckoo hashing scheme with $\eta$ hash functions, each inserted element, say $u$, could possibly reside in at most $\eta$ bins at $T_\mathsf{cuckoo}$.
For each $\mathcal{C}_i$ with distinct selection $\textbf{s}^{(i)}$, the exact position of $u$ looks random to whoever does not know $\textbf{s}^{(i)}$.
With such property, the client can obliviously instruct servers to aggregate his weight updates.
For instance, if we denote the possible location of $u$ as $h_1(u),\ldots,h_\eta(u)$, their distinct positions as $\mathsf{pos}_{h_1(u)}, \ldots, \mathsf{pos}_{h_\eta(u)}$, and the DPF keys for all locations as $k_b[h_1(u)], \ldots, k_b[h_\eta(u)]$, each $\mathcal{S}_b$ can locally compute $\sum_{d=1}^\eta\mathsf{DPF.Eval}(k_b[h_d(u)], \mathsf{pos}_{h_d(u)})$ and denote the result as $[\Delta w^{(i)}_u]_b$.
The correctness can also be easily verified by the following,

\begin{align*}
    \Delta w^{(i)}_u=[\Delta w^{(i)}_u]_0 + &[\Delta w^{(i)}_u]_1\in\GG\\
    &=\sum_d^\eta f_{\mathsf{pos}_{h_d(u)}, *}(\mathsf{pos}_{h_\eta(u)})\in\GG
\end{align*}
As for a concrete example, in Figure \ref{fig:cuckoo}, a client has an index set $\set{1,4}$ and the client inserts its index set into the cuckoo table.
Here, we also construct the simple table with the full index set $\set{1,\ldots, 5}$.
Note that the element ``4'' is inserted into both Bin-1 and Bin-3.
Therefore, to aggregate $\Delta w^{(i)}_4$, $\mathcal{S}_b$ invokes $[\Delta w^{(i)}_4]_b\gets\mathsf{DPF.Eval}(k_b[1], 2) + \mathsf{DPF.Eval}(k_b[3], 2)$ as we have defined before, and since only Bin-1's evaluation is correct, both Bin-2 and Bin-3 evaluates to zero, we can easily verify the correctness of our protocols.

\medskip
\noindent
\textbf{Master seed for each client.}
Considering semi-honest adversaries, we observe that each client could use a private master seed and a public pseudorandom function (PRF) with a public number to generate independently random DPF seeds in the random oracle model.
Instead of sending a seed for each bin, now each client can send one master seed and instruct servers to expand it.
For instance, each client samples two master keys $\mathsf{msk}_0, \mathsf{msk}_1\sample\bin^\lambda$, then invokes $\prf(\mathsf{msk}_b, i)$ to generate $i$-th bin's DPF master seed for server $b$.
This optimisation technique essentially reduces the client-side communication from sending $B$ (the bin number, i.e. $B=\epsilon k$) seeds to one master key for each server.

\medskip
\noindent
\textbf{Handling dummy bins.}
Moreover, cuckoo hashing generates $\epsilon k$ bins with $\epsilon>1$ when inserting $k$ elements, we let each client generates dummy DPF keys for the empty bins.
Those dummy DPF keys preserve the structure of normal DPF keys except they evaluate to zero on all possible inputs.
One can instantiate the generation of those keys by $\mathsf{DPF.Gen}(1^\lambda, 0, 0)$ without loss of generality.

\begin{figure}[t!]
    \centering
    \includegraphics[width=0.75\linewidth]{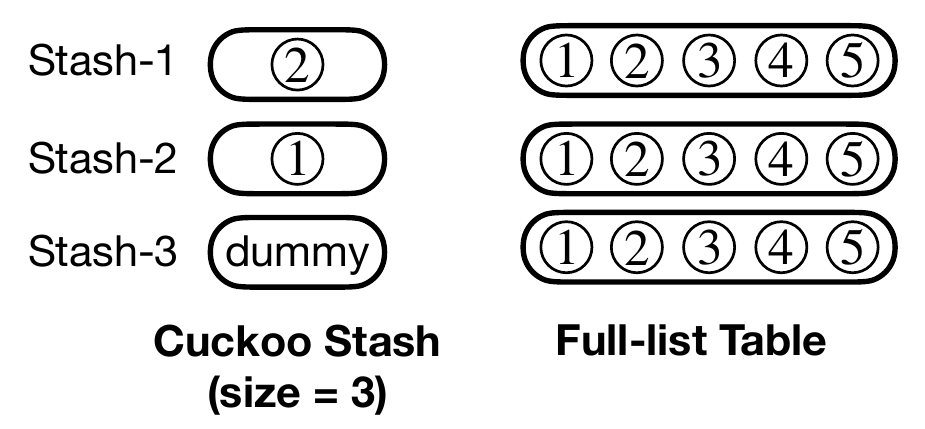}
    \caption{An example: handling cuckoo hash stashes. This is the case where the whole set is $\set{1,\ldots, 5}$, and element $1,2$ is inserted into cuckoo stash. Note that we preset cuckoo hash stash size to $3$ which is data independent.}
    \label{fig:stash}
\end{figure}

\medskip
\noindent
\textbf{Handling stash.}
Previously, we only introduce our basic protocols in the stash-less setting (i.e. stash size = 0).
Now, we show how each $\mathcal{C}_i$ handles the stash.
Assume each $\mathcal{C}_i$ has $\mathsf{stash}^{(i)}$ with stash size of $\sigma>0$.
Then for each stash element $\mathsf{stash}^{(i)}[j]$, client generates DPF keys for the input range of the full model index.
Now, each server evaluates and aggregates bin $h_1(u), \ldots, h_\eta(u)$ for element $u$, additionally with evaluating $u$ on all stash keys.
For the same example, if we have the stash in Figure \ref{fig:stash}, to aggregate $\Delta w^{(i)}_4$ $\mathcal{S}_b$ invokes:
\begin{align*}
    [\Delta w^{(i)}_4]_b\gets\mathsf{DPF.Eval}(k_b[1], 2) + \mathsf{DPF.Eval}(k_b[3], 2)\\
    + \sum_{d=1}^\sigma\mathsf{DPF.Eval}(\hat{k}_b[d], 4)\in\GG
\end{align*}
where we denote the stash keys as $\hat{k}_b[1], \hat{k}_b[2], \hat{k}_b[3]$.



\begin{lemma}
    The basic PSR protocol and SSA protocol in \ref{fig:protocol} are correct instantiations of their functionalities and are secure against arbitrarily semi-honest adversaries with an honest server.
\end{lemma}



%
%

Without loss of generality, we treat each stash element as additional cuckoo bins in both PSR and SSA protocols, where the corresponding simple bins each contains $\set{1,\ldots, m}$.
To verify the correctness of PSR, for $1\leq j\leq B+\sigma$, we have
\begin{align*}
    [w_j^{(i)}] = & \sum_{d=1}^{|T_\mathsf{simple}[j]|}w_{T_\mathsf{simple}[j][d]}\cdot \\
    &\left(\mathsf{DPF.Eval}(k_0^{(i)}[j], d) +  \mathsf{DPF.Eval}(k_1^{(i)}[j], d)\in\GG\right)
\end{align*}
where for the case of $j>B$, we define $k_b^{(i)}[j]=\hat{k}_b^{(i)}[j-B]$.
Assume the DPF is a correct instantiation of point function, for the $j$-th bin, its two DPF keys jointly implies $f_{\mathsf{pos}_j\to 1}$.
To this end, cuckoo hashing ensures that only one position (i.e. $\mathsf{pos}_j$) is selected by client for each bin.
By aggregating all evaluation results where there is only one non-zero result, we have $w_j^{(i)}=w_{T_\mathsf{simple}[j][\mathsf{pos}_j]}$, which equals to $w_{T_\mathsf{cuckoo}[j]}$ at the $j$-th bin.
Note that $\mathsf{pos}_j$ indicates the position of an element at $T_\mathsf{simple}[j]$ that the client desires.




%
Similarly, to verify the correctness of SSA, we evaluate all positions at all bins.
For $1\leq j\leq m$, 
\begin{align*}
    [\Delta w_j]=\sum_{i=1}^n \sum_{d=1}^\eta
    \Big(&\mathsf{DPF.Eval}(k_0^{(i)}[h_d(j)], \mathsf{pos}_{h_d(j)}) + \\ 
    &\mathsf{DPF.Eval}(k_1^{(i)}[h_d(j)], \mathsf{pos}_{h_d(j)})\Big)\\
    + \sum_{t=1}^\sigma \Big(&\mathsf{DPF.Eval}(\hat{k}_0^{(i)}[t], j) +\\
    &\mathsf{DPF.Eval}(\hat{k}_1^{(i)}[t], j)\Big)\in\GG
\end{align*}
To verify the correctness, for each client $i$, first, find all possible bins and positions that each $\Delta w_j$ may reside in, then evaluates and aggregates all bins and positions.
If $j\notin s^{(i)}$, the evaluation result equals to zero, and otherwise the evaluation result equals to $\Delta w_j^{(i)}$.
Again since each client chooses only one bin with one position, we have $\Delta w_j=\sum_{i=1}^n\Delta w_j^{(i)}$.

\medskip
\noindent
\textbf{Efficiency}.
Intuitively, each client in our semi-honest protocol uploads $2\epsilon k$ DPF keys.
\textit{Without loss of generality, we can treat our basic PSR as a special SSA protocol with $\GG=\bin$.}
Observe that our DPF construction~\cite{boyle2016function} has a public part (which is the same for both keys) with $\ceil{\log\Theta}(\lambda+2) + \ceil{\log|\GG|}$ bit, and a private part (which differs between the two keys) with $\lambda$ bit.
Each client can upload the public parts to one server, and then the private part to each server, respectively.
Recall that, with random oracle assumption, we could use master seed to generate DPF initial seeds for each bin, (for the stash-less setting) the upload communication for each client reduces to
\begin{align*}
    \epsilon k \left( \ceil{\log\Theta}(\lambda+2) + \ceil{\log|\GG|}\right)  + \lambda.
\end{align*}

\smallskip
\noindent
\textbf{Limitations of the basic SSA protocol}
Recall the trivial protocol has communication of $m\cdot\ceil{\log|\GG|} + \lambda$ bits.
If we let the plaintext compression rate be $k/m=c$, our protocol is non-trivial conditioned on the following,
\begin{align*}
    \epsilon k \left( \ceil{\log\Theta}(\lambda+2) + \ceil{\log|\GG|}\right)  + \lambda - (m\ceil{\log|\GG|} + \lambda) &> 0\\
    \epsilon k (\lambda+2)\ceil{\log\Theta} - (m-\epsilon k)\ceil{\log|\GG|} &> 0.
\end{align*}
If we assume $\lambda=128$, $\epsilon=1.25$, $\ceil{\log\GG}=128$ and $\ceil{\log\Theta}=9$, then we have $1622.5 k > 128m$, which meas that our basic SSA protocol is non-trivial only when we have $c\leq 7.8\%$.
Note that our basic PSR is already a non-trivial PIR protocol, therefore we ignore the proof of its non-triviality.
To deal with this limitation, in section \ref{sec:optimisations} we further propose several optimisations for different scenarios.

\begin{figure*}[hbtp]
    \centering
    \input{protocols/basic}
    \caption{Basic Private Submodel Retrieval (PSR) and Secure Submodel Aggregation (SSA) Protocols (semi-honest version)}
    \label{fig:protocol}
\end{figure*}

\section{Updatable Distributed Point Function}
\label{sec:updatable-distributed-point-function}

Now, we introduce a new tool called updatable distributed point function, which allows the client to obliviously update DPF keys from $f_{\alpha,\beta}$ to $f_{\alpha, \beta^\prime}$.
More formally, UDPF includes the following algorithms:
\begin{description}
    \item[$\mathsf{Gen}(1^\lambda, \alpha, \beta):$] Given security parameter $\lambda\in\NN$, a point $\alpha\in\bin^n$ and a value $\beta\in\GG$, output two DPF keys $(k_0, k_1)$.
    \item[$\mathsf{Eval}(b, k_b, x, e):$] Given a DPF key $k_b$, a point $x\in\bin^n$ for epoch $e$, output the secret-shared evaluation result.
    \item[$\mathsf{Next}(k_0, k_1, \beta^\prime, e)$:] Takes input as a pair of existing DPF keys $(k_0, k_1)$, a new $\beta^\prime\in\GG$ for epoch $e$, outputs a hint.
    \item[$\mathsf{Update}(k_b, \mathsf{hint}, e)$:] Takes input as a DPF key $k_b$ where $b\in\bin$, and a hint for epoch $e$, outputs a new DPF key.
\end{description}

\smallskip
\noindent
\textbf{Correctness.}
The correctness of the UDPF requires that the update of a valid pair of DPF keys $k_0, k_1$ (which implies $f_{\alpha, \beta}$) from epoch $e$ to $e+1$ with $\beta^\prime$ leads again to a pair of valid DPF keys $k_0^\prime, k_1^\prime$ (which implies $f_{\alpha, \beta^\prime}$).

\smallskip
\noindent
\textbf{Security.}
Intuitively, we want to say that a server who holds $k_b, k^\prime_b$ learns nothing about $\alpha, \beta, \beta^\prime$.
We use indistinguishability based security definition, where given $\alpha, \beta$, first calculate $(k_0, k_1)\gets \mathsf{Gen}(1^\lambda, \alpha, \beta)$, then for all $\beta^\prime$ at epoch $e$, the following holds,
\begin{align*}
    \{k_b^\prime|(k_0^\prime, k_1^\prime)&\gets\mathsf{DPF.Next}(k_0, k_1, \beta^\prime, e)\}_{k_0, k_1, b}\\
    \cindist
    \{k_b^\ast|(k_0^\ast, k_1^\ast)&\gets\mathsf{DPF.Next}(k_0, k_1, \beta_r, e)\}_{k_0, k_1, b}
\end{align*}

Recall the security of standard DPF \cite{boyle2015function} says that for all $\lambda\in\NN, \alpha, \beta, b\in\bin$, the following holds,
\begin{align*}
    \{k_b|(k_0, k_1)\gets\mathsf{Gen}(1^\lambda, \alpha, \beta)\}
    \cindist\{\simulator(1^\lambda, \GG, n)\}
\end{align*}

\smallskip
\noindent
\textbf{Construction.}
We would like to let client to obliviously send a \textit{short} hint to each server, and each server updates its DPF key without learning anything else.
One trivial answer is to let client re-run $\mathsf{DPF.Gen}$ for $\alpha$ and $\beta^\prime$, but it would generate the hint with the same size as DPF key.
Recall that in the original construction of DPF \cite{boyle2015function}, $\beta$ is hidden in the final correlation word.
Therefore, another idea is to replace the last correlation word, i.e. $CW^{(n+1)}$ in each DPF key.
Now that we only replace the last correlation word of the original key, we can simplify the above security definition to:
\begin{align*}
    \{(CW^{(n+1)})^\prime\in k_b^\prime\}\cindist \{(CW^{(n+1)})^\ast\in k_b^\ast\}.
\end{align*}
Previous standard construction of $CW^{(n+1)}$ fails to meet the security definition, where
\begin{align*}
    CW^{(n+1)}\gets(-1)^{t_1^{(n)}}\cdot[\beta-\prg(s_0^{(n)}) + \prg(s_1^{(n)})]\in\GG.
\end{align*}
Assume the existence of a random oracle $\hash: \bin^\lambda\times\NN \to\GG$, the following $CW^{(n+1)}$ meets the security definition.
\begin{align*}
    CW^{(n+1)}\gets(-1)^{t_1^{(n)}}\cdot[\beta-\hash(s_0^{(n)}, e) + \hash(s_1^{(n)}, e)]\in\GG.
\end{align*}

\section{Optimisations}
\label{sec:optimisations}

In the previous section, we explain in concept how to solve the submodel update problem by leveraging DPF and Cuckoo hashing.
And we also introduced a new tool updatable DPF to further improve protocol efficiency.
However, there are still fundamental limitations of our basic protocols.
Notice that in the following we do not discuss the efficiency of our basic PSR protocol as it is state-of-art two-server PIR protocol.
We mainly focus on the potential further optimisations for our basic SSA protocol to improve its efficiency and applicability for real-world applications.

\medskip
\noindent
\textbf{Limitations of the basic protocols}.
Though the basic protocols are both correct and secure, we still need to show that they are non-trivial.
In our semi-honest protocol with stash-less setting, at each communication round, each client sends concretely $\epsilon k$ DPF keys to one server, and a master seed to the other.
Let $l=\ceil{\log(|\GG|)}=128$ bit to represent a weight, and security parameter $\lambda=128$ (as recommended by NIST \cite{barker2016nist}).
Also let cuckoo parameters $\epsilon\approx1.25$, and $\ceil{\log\Theta}=9$.
If we define $c=k/m$ as the compression rate, where $m$ is the total number of weights, we can calculate the communication advantage rate for our basic SSA protocol, $\mathcal{R}(\pi_\mathsf{ssa})\approx 12.68 c$.
In another word, it is non-trivial only when $c\lessapprox 7.8\%$. 


\medskip
\noindent
In addition to the basic protocols, we propose optimisations specially designed for different scenarios.
We present a short summary of the optimisation techniques in Table \ref{tab:optimisation}.

\begin{table}[t!]
    \centering
    \begin{tabular}{@{}ll@{}}
    \toprule
    \textbf{Scenario} & \textbf{Techniques}\\
    \midrule
    $k \ll m$ & Basic protocol\\
    $|\bigcup_i s^{(i)}| \ll m$ & Basic protocol + \emph{PSU}\\
    Fixed submodel & Basic protocol + \emph{U-DPF}\\
    Allow grouping top-$k$ & Basic protocol + \emph{Mega-Element}\\
    \bottomrule
    \end{tabular}
    \caption{A summary of optimisation techniques}
    \label{tab:optimisation}
\end{table}

\medskip
\noindent
\textbf{Basic protocol with Private Set Union (PSU).}
The baseline SFL trivially uses the prohibitively large full index set (i.e., the position of the full model) for alignment. 
As is observed by \cite{niu2020billion}, we could use the union of the chosen clients’ real index sets as the  necessary and sufficient scope for alignment.
Because the union is much smaller than the full index set in certain scenarios, secure FSL can significantly outperform secure FL without sacrificing privacy.
Recall that in FSL framework, process \textcircled{\raisebox{-1.5pt}{1}} selects a certain number of clients from all client candidates.
The union set of client's selection, namely $\bigcup_i s^{(i)}$ could be a small subset of $\{1,...,m\}$.
Here, assume the leakage of the union set reveals negligible useful information about the clients, we can include a \emph{Private Set Union} (PSU) protocol after process \textcircled{\raisebox{-1.5pt}{2}} (local train).
The union result is then revealed to all participants, including two servers.
PSU protocol ensures that clients only learn their union set, nothing else.
Finally, all parties invoke $t$-out-of-$\bigcup_i s^{(i)}$ secure aggregation protocol.
Specifically, this optimisation reduces the maximum size of bins in $T_\mathsf{simple}$, namely $\Theta$.
For example, in our experiments, PSU decreases $\Theta$ from $9$ to $5$.
In this case, we have $\mathcal{R}(\pi_\mathsf{ssa})\leq 1$ when $c\lessapprox 13.4\%$.

There are many existing efficient PSU protocols~\cite{Kolesnikov2019ScalablePS}.
Note that this optimisation is only non-trivial if the saved cost is larger than the communication overhead brought by PSU.
We observe that exiting PSU protocols have already achieve remarkable efficiency, and therefore this optimisation could be considered as non-trivial.
In addtion, we believe that the construction of PSU is of independent interest.

\medskip
\noindent
\textbf{Basic protocol with Updatable DPF.}
In federated learning systems, personalization techniques, such as multi-task learning or transfer learning \cite{kulkarni2020survey}, are widely used for optimizing the model performances for heterogeneous clients, where clients may vary in data distribution, local tasks, or even different computation and communication capabilities.
In such cases, clients may possess different submodels or even compound models that are relevant to them by design. 
For example, recent work proposes a federated learning framework called \emph{HeteroFL} to train heterogeneous local models, while still a single global model is aggregated for inference~\cite{diao2020heterofl}, experimenting with the cases of both fixed and dynamic submodels. 
A typical training process for fixed submodels is the same as the dynamic case, that is, to upload the relevant submodels and download the whole model. 
The same question may arise: how can we achieve communication efficient yet secure model aggregation, especially when the submodels are sparse ( i.e., entries corresponding to a sub-model only contribute to a small portion of the full model).

As for the case of fixed submodel, where we assume each client queries the same $\textbf{s}^{(i)}$ for each training round, we could leverages Updatable DPF to further improve the protocol efficiency.
Recall that U-DPF allows client to update the DPF keys such that it implies the same point but different value.
Considering the case of fixed submodel, we can let client generate the DPF key in the first round, and then update the DPF keys in all rest rounds.
The key observation is that each client could update DPF keys with communication size only equal to the hint (which is generated by algorithm $\mathsf{GenHint}$).
In particular for a training task, the communication of the first round is identical to the basic SSA protocol while the communication of other rounds is exactly the size of hint, aka, $k\cdot l$ bits.
This brings the rate of our protocol $\pi$ to $\mathcal{R}^{(1)}(\pi_\mathsf{udpf})=\mathcal{R}(\pi_\mathsf{ssa})$, and $\mathcal{R}^{(>1)}(\pi_\mathsf{udpf})=c$.

\begin{figure}[t!]
    \centering
    \includegraphics[width=0.5\linewidth]{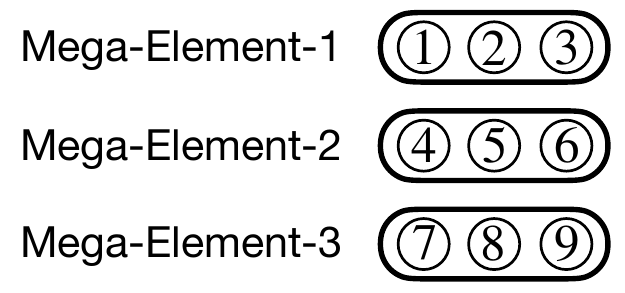}
    \caption{An example of mega-elements. In this example, we group elements $\set{1,\ldots, 9}$ into three mega-elements, where each mega-element has the size of $3$.}
    \label{fig:mega-element}
\end{figure}

\medskip
\noindent
\textbf{Basic protocol with Mega-Element.}
As is illustrated above, the protocol overhead $\mathcal{R}(\pi_\mathsf{ssa})$ is dominated by the size of weight, i.e. $l$. The overhead can be reduced by increasing $l$, either in terms of its representation size (rather impractical) or grouping the weights as mega-elements (shown in Fig. \ref{fig:mega-element}).
The size of the ``mega-element" is equivalent to the concept of the original weight element, where we denote the mega-element size as $L = \tau l$.
In this sense, our submodel aggregation protocol manipulates on the basis of mega-elements, i.e., parameter groups. 
\begin{align}
   \mathcal{R}(\pi_\mathsf{mega\text{-}elem})=c\cdot \frac{\epsilon ((\lambda +2) \ceil{\log\Theta} + L)}{\tau l}.
   \label{eq:rate}
\end{align}

Fortunately, such structured parameter grouping patterns naturally exist in many widely used deep learning models, such as embedding models.
To be more specific, embedding models, such as language models (e.g., Word2Vec~\cite{mikolov2013distributed}, TextCNN~\cite{Kim2014ConvolutionalNN},
ELMo~\cite{peters2018deep}, BERT~\cite{devlin2018bert}, etc.) or deep recommendation models (e.g., PNN~\cite{qu2016product}, PinSage~\cite{ying2018graph}, DIN~\cite{zhou2018deep}, etc.), operate on top of the embedding layer with each row (i.e., an embedding vector in the embedding matrix) corresponding to a specific word in the vocabulary used for a language model or a certain item for the recommendation. Embedding models have revolutionized the way to represent the text/items and have become the dominant approach for NLP and recommendation tasks.
%
As pointed in the prior work~\cite{niu2020billion}, Taobao.com\footnote{One of the biggest e-commerce platforms in China.} has approximately 2 billion items, thus, the recommendation model trained is with 98.22\% of model parameters in the embedding layer, where each item is embedded into a vector of dimension size as 18. 
In this case, if we group the 18 elements as a mega-element (i.e., $\tau=18$), for Eq. (\ref{eq:rate}), we can have $\mathcal{R}(\pi_\mathsf{mega\text{-}elem})\leq 1$ when $c\lessapprox 53.1\%$, not to mention the embedding dimension is typically a two-digit or even three-digit number.
Note that we let $\epsilon=1.25$, $l=\lambda=128$, $\ceil{\log\Theta}=9$.
In this way, the compression rate can be dramatically reduced for secure FSL together with the Mega-Element grouping strategy.

\section{Empirical Experiments}
\label{sec:experiments}

Our experiments tend to answer the following questions.
(1) How should we choose cuckoo hashing parameters?
(2) How efficient is our basic protocols, in terms of communication and runtime, when considering different weight sizes and different submodel sizes?
(3) How practical is our basic protocol on a regular machine learning task?
(4) Is our protocol more efficient than the state-of-art full-blown secure aggregation protocol?
In our experiments, we use the \textit{top-$k$ sparsification strategy} \cite{aji2017sparse} for submodel selection, which is widely used for learning robust and communication-efficient federated models with heterogeneous clients.

\medskip
\noindent
\textbf{Implementation and setup.}
We run all of our efficiency experiments on a machine with 64 Core 2.5GHz Intel(R) Xeon(R) CPU with 188G memory, we compile our program using modern C++ compiler (with support for C++ standard 14), and packed our code into a python library.
In addition, our tests were run in a local network, with $\approx$ 3ms network latency.
Also, we use 128-bit computational security parameter and 40-bit statistical security parameter.
Similarly, we let hash failure probability to be at most $\leq 2^{-40}$.

\subsection{Parameter Choices}

We run our experiment in the stash-less setting, where cuckoo stash size $\sigma=0$, and hash number $\eta=3$.

\medskip\noindent\textbf{Scale factor}.
We run our experiments on different input size, i.e., $\set{2^{10}, 2^{15}, 2^{20}, 2^{25}}$, while ensuring that hash failure probability to be less than $2^{-40}$.

\begin{table}[t]
    \centering
    \begin{tabular}{@{}ccccc@{}}
    \toprule
    \textbf{Input size}   &  $2^{10}$ & $2^{15}$ & $2^{20}$ & $2^{25}$  \\
    \midrule
    $\epsilon$ & 1.25 & 1.25 & 1.27 & 1.28\\
    \bottomrule
    \end{tabular}
    \caption{Scale factor choice}
    \label{tab:scale-factor}
\end{table}

\medskip
\noindent
\textbf{Maximum bin size of $T_\mathsf{simple}$}.
We run our experiments using simple hashing and inserts incremental index sets.
For instance, for a set size of $2^{10}$, we insert the set $\set{1,...,2^{10}}$ to $T_\mathsf{simple}$ with the above cuckoo parameter, and then check the maximum bin size in $T_\mathsf{simple}$.
We show our experiments result in table \ref{tab:max-bin-size}.
It shows that with item number less than $2^{25}$, and compression rate more than $1\%$, it is practically sufficient to set a fixed $\ceil{\log\Theta}=9$ for the DPF key generation scheme.
In the following experiments, we use adaptive $\Theta$ for each bin, we only use fixed $\ceil{\log\Theta}=9$ to obtain an approximate amount of communication.
\begin{table}[t]
    \centering
    \begin{tabular}{@{}ccccc@{}}
    \toprule
         &  $2^{10}$ & $2^{15}$ & $2^{20}$ & $2^{25}$  \\
    \midrule
    $1\%$ & 324 & 315 & 336 & \textbf{366}\\
    $10\%$ & 45 & 54 & 66 & 78\\
    $30\%$ & 27 & 36 & 39 & 48\\
    $50\%$ & 21 & 24 & 30 & 36\\
    $70\%$ & 18 & 21 & 27 & 30\\
    \bottomrule
    \end{tabular}
    \caption{Maximum bin size with different compression rate and weight size}
    \label{tab:max-bin-size}
\end{table}

\begin{figure*}[t!]
    \centering
    \begin{subfigure}[b]{0.32\linewidth}
        \centering
        \includegraphics[width=\linewidth]{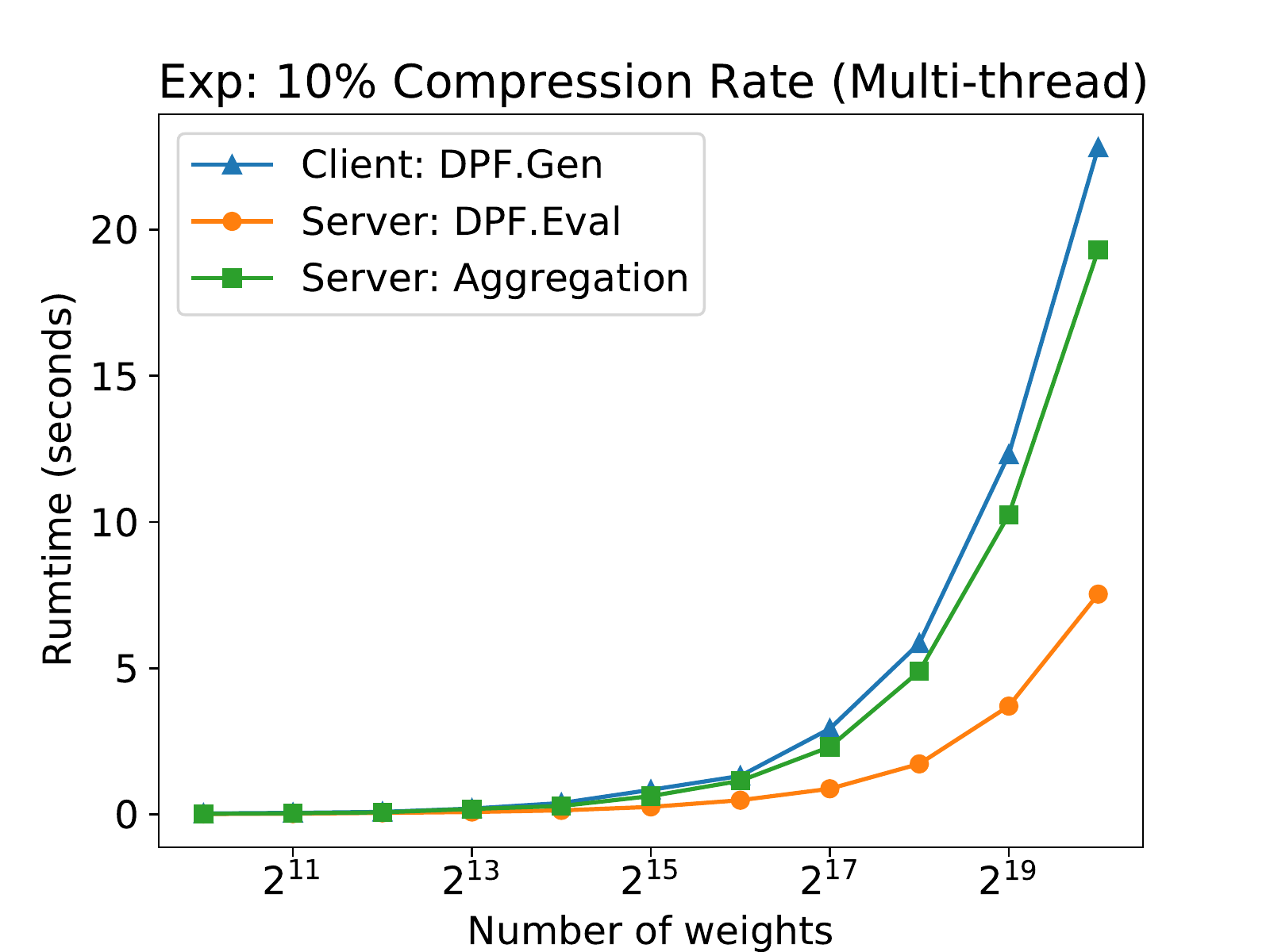}
        \caption{}
        \label{fig:exp01-10}
    \end{subfigure}
    \hfill
    \begin{subfigure}[b]{0.32\linewidth}
        \centering
        \includegraphics[width=\linewidth]{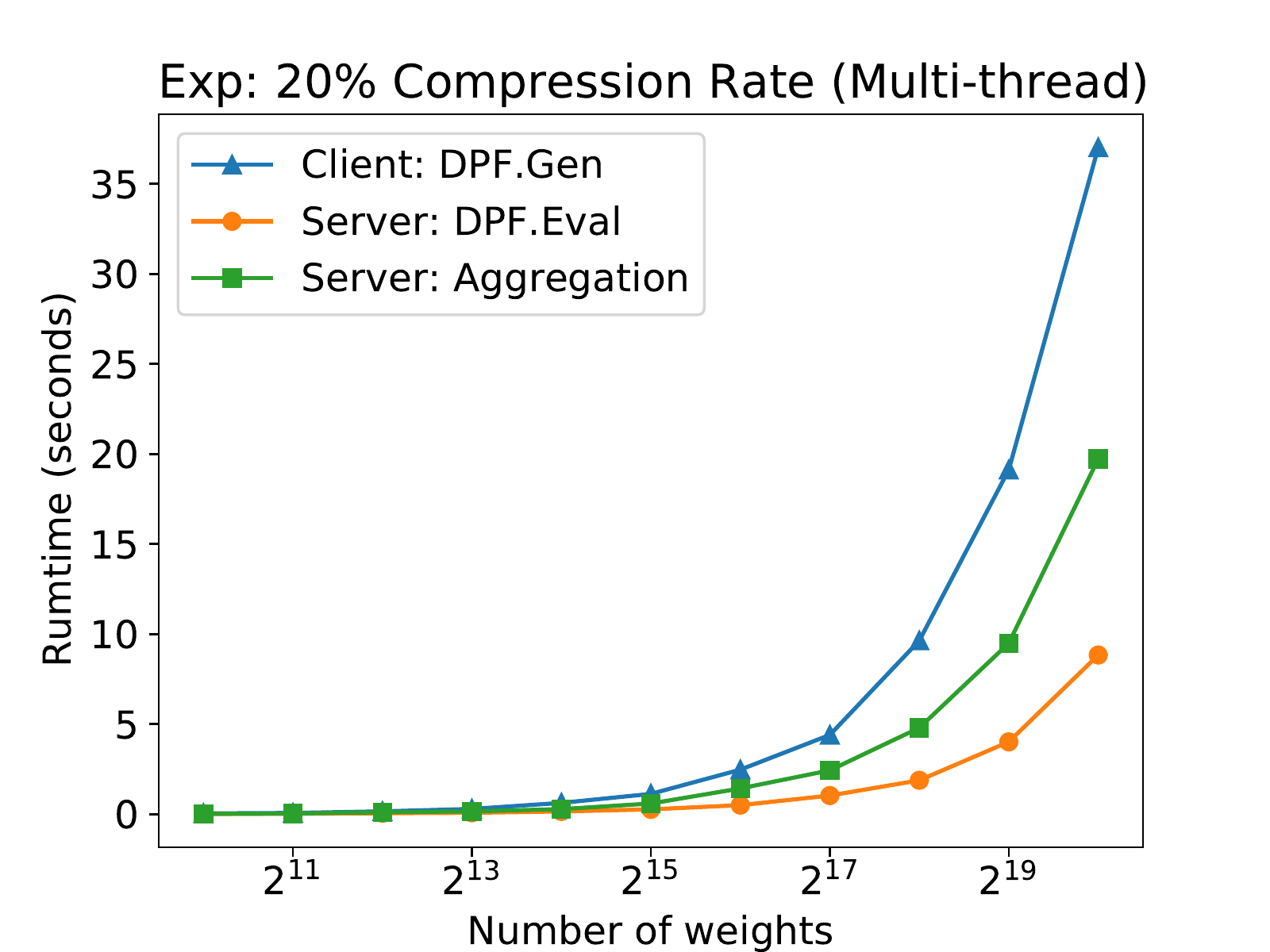}
        \caption{}
        \label{fig:exp01-20}
    \end{subfigure}
    \hfill
    \begin{subfigure}[b]{0.32\linewidth}
        \centering
        \includegraphics[width=\linewidth]{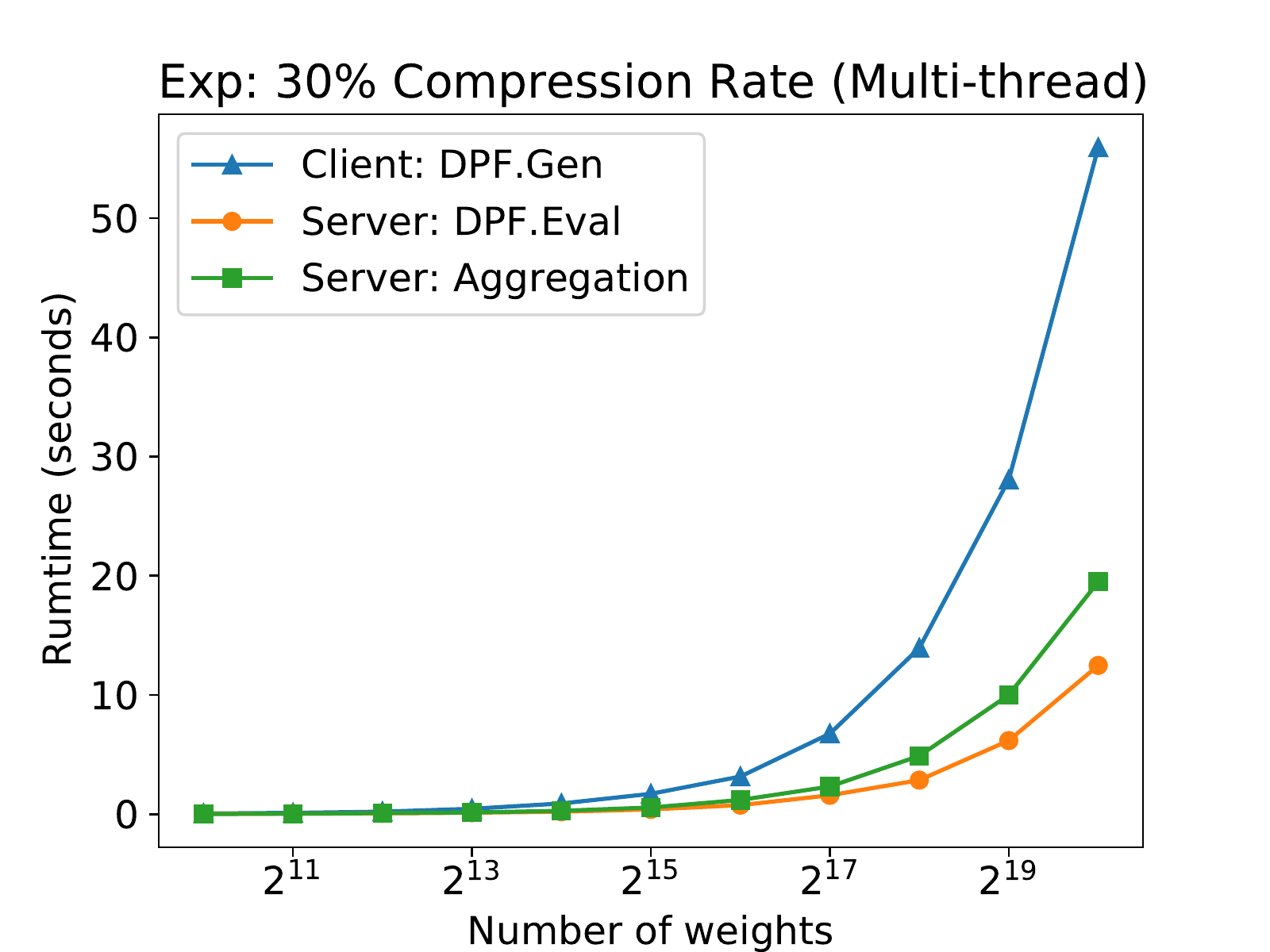}
        \caption{}
        \label{fig:exp01-30}
    \end{subfigure}
    \caption{Computation efficiency of our basic SSA protocol with various number of weights and $10\%\sim30\%$ compression rates.}
    \label{fig:exp01}
\end{figure*}

\begin{table*}[t!]
    \centering
    \begin{tabular}{@{}l cccccccccccc@{}}
    \toprule
    $m$     &&  \multicolumn{3}{c}{$2^{10}$} && \multicolumn{3}{c}{$2^{15}$} && \multicolumn{3}{c}{$2^{20}$} \\
    \cmidrule{3-5} \cmidrule{7-9} \cmidrule{11-13}
    $k/m$   &&   10\% & 20\% & 30\% && 10\% & 20\% & 30\% && 10\% & 20\% & 30\% \\
    \midrule
    DPF Gen time                && 0.028s & 0.045s & 0.056s && 0.838s & 1.129s & 1.707s && 22.806s & 37.018s & 55.918s\\
    DPF Eval time               && 0.012s & 0.015s & 7.532s && 0.253s & 0.124s & 0.197s && 7.532s & 0.98s & 1.732s\\
    Aggregation time                 && 0.023s & 0.018s & 0.018s && 0.018s & 0.179s & 0.172s && 0.018s & 1.844s & 2.257s\\
    \bottomrule
    \end{tabular}
    \caption{Computation efficiency of our basic SSA protocol (in seconds), where we let $l=128$, $\lambda=128$, $\ceil{\log\Theta}=9$, and see table \ref{tab:scale-factor} for $\epsilon$ choices. Note that each client only invokes DPF Gen in our protocol while each server invokes both DPF Eval and Aggregation in our protocol. The aggregation refers to the process that the server locally aggregates all possible evaluation results for each output.}
    \label{tab:basic-comp}
\end{table*}

\begin{table*}[t!]
    \centering
    \begin{tabular}{@{}l cccccccccccc@{}}
    \toprule
    $m$     &&  \multicolumn{3}{c}{$2^{10}$} && \multicolumn{3}{c}{$2^{15}$} && \multicolumn{3}{c}{$2^{20}$} \\
    \cmidrule{0-0} \cmidrule{3-5} \cmidrule{7-9} \cmidrule{11-13}
    Secure Aggregation && \multicolumn{3}{c}{0.015} && \multicolumn{3}{c}{0.5} && \multicolumn{3}{c}{16}\\
    \scriptsize{(compression rate $c=k/m$)} &&  {} & \scriptsize{(100\%)} & {} && {} & \scriptsize{(100\%)} & {} && {} & \scriptsize{(100\%)} & {}\\
    Our protocol  && 0.002 & 0.009 & 0.019 && 0.063 & 0.317 & 0.633 && 2.028 & 10.14 & 20.28\\
    \scriptsize{(compression rate $c=k/m$)} &&   \scriptsize{(1\%)} & \scriptsize{(5\%)} & \scriptsize{(10\%)} && \scriptsize{(1\%)} & \scriptsize{(5\%)} & \scriptsize{(10\%)} && \scriptsize{(1\%)} & \scriptsize{(5\%)} & \scriptsize{(10\%)}\\
    \bottomrule
    \end{tabular}
    \caption{Communication efficiency of our basic SSA protocol (in MB), where we let $l=128$, $\lambda=128$, $\ceil{\log\Theta}=9$, and see table \ref{tab:scale-factor} for $\epsilon$ choices. Note that the secure aggregation we refer to in the table is the na\"ive secure aggregation protocol in the two-server setting.}
    \label{tab:basic-comm}
\end{table*}

\begin{figure}[t!]
    \centering
    \includegraphics[width=0.95\linewidth]{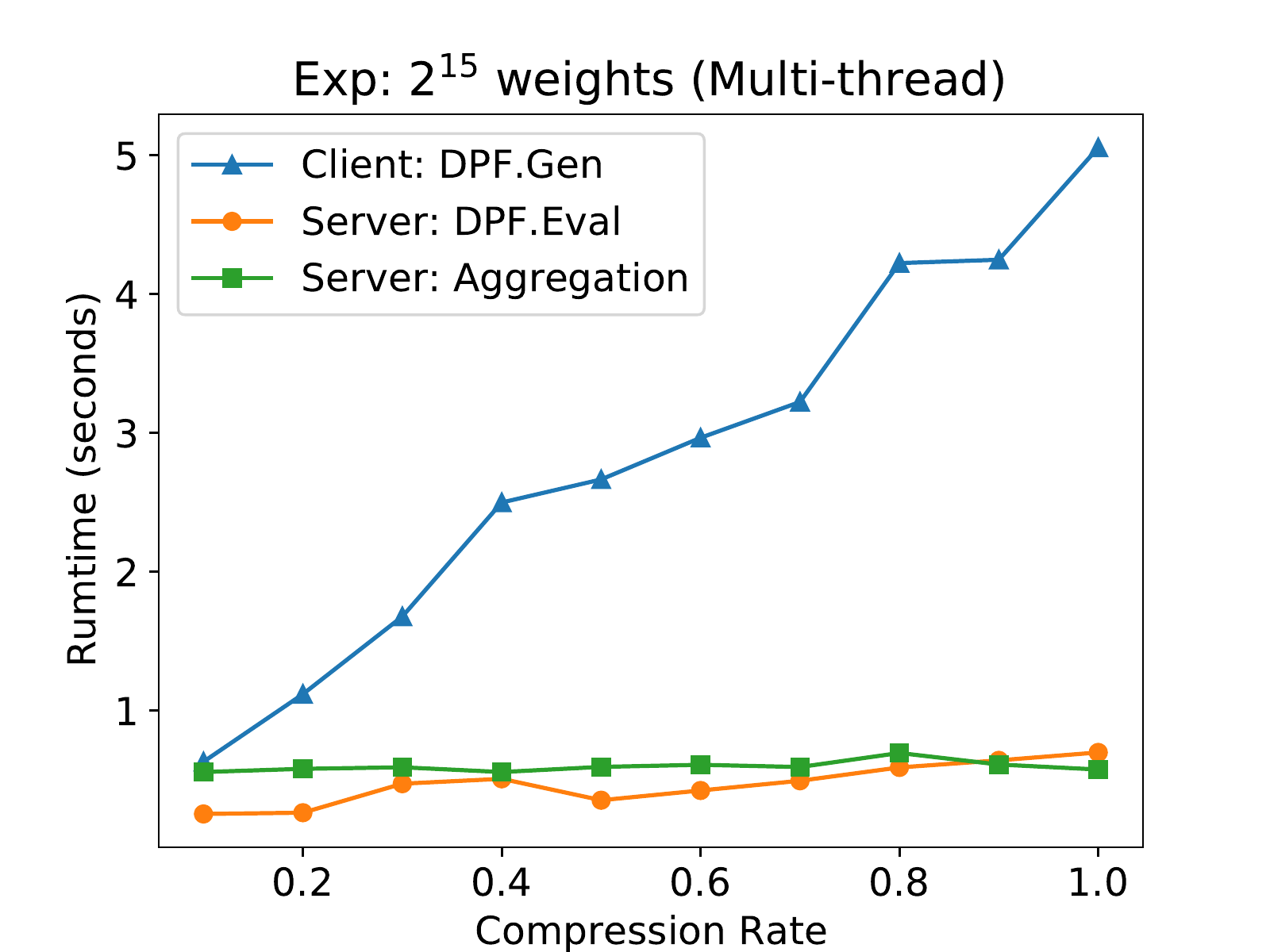}
    \caption{Protocol efficiency experiments 2.}
    \label{fig:exp02}
\end{figure}

\subsection{Protocol Efficiency}

Recall that we provide theoretical analysis of the communication efficiency of our basic protocols, and with optimisations.
In this section, we focus on the computational efficiency.
First, we adopt several optimisations from previous work, i.e.,
enabling full-domain evaluation of DPF \cite{boyle2016function} and we enable multi-threading for all our experiments.
We run our experiments with different input sizes (from $2^{10}$ to $2^{20}$) with different compression rate, i.e., $10\%, 20\%, 30\%$.
We choose $10\%\sim30\%$ compression rate because we observe that this compression range is acceptable for most federated learning tasks.
See Figure \ref{fig:exp01}.
We record the the DPF key generation time for one client, and the DPF evaluation and aggregation time for each server.
All our experiments can finish within 30s when there is less than 33 million weights with $10\%$ compression rate.
Second, we run our experiments with $2^{15}$ weights and different compression rates, ranging from $10\%$ to $100\%$.
See Figure \ref{fig:exp02} for experimental results.
For more general computational efficiency results, see Table \ref{tab:basic-comp}.
Note that the server computation runtime is almost irrelevant to the compression rate, while the client runtime is linear to the compression rate.
In terms of communication efficiency, we run our experiments with different input sizes, ranging from $2^{10}$ to $2^{20}$, and different compression rates, i.e., $1\%\sim10\%$ (since our basic protocol is only non-trivial in this range without scenario-specific optimisations).
See Table \ref{tab:basic-comm} for detailed results.

\begin{table*}[t!]
    \centering
    \begin{tabular}{@{}l cccccccccc@{}}
    \toprule
    Dataset   & 5\% & 10\% & 20\% & 40\% & 60\% & 80\% & 100\% \\
    \midrule
    MNIST  & 97.36 $\pm$ 0.12 & 97.40 $\pm$ 0.13 & 97.43 $\pm$ 0.13 & 97.45 $\pm$ 0.13 & 97.47 $\pm$ 0.13 & 97.47 $\pm$ 0.13 & 97.47 $\pm$ 0.13\\
    CIFAR10 & 58.79 $\pm$ 1.17 & 59.15 $\pm$ 1.17 & 59.46 $\pm$ 1.09 & 59.58 $\pm$ 1.05 & 59.62 $\pm$ 1.02 & 59.67 $\pm$ 1.12 & 59.57 $\pm$ 1.10\\
    TREC(iid) & 88.87 $\pm$ 0.50 & 89.20 $\pm$ 0.87 & 89.20 $\pm$ 0.20 & 89.47 $\pm$ 0.70 & 89.60 $\pm$ 0.87 & 89.60 $\pm$ 0.87 & 89.60 $\pm$ 0.87 \\
    \bottomrule
    \end{tabular}
    \caption{Accuracy Experiment 1 (Basic Protocol) for real-world FSL tasks}
    \label{tab:naive-topk}
\end{table*}

\begin{table*}[t!]
    \centering
    \begin{tabular}{@{}l cccccccccc@{}}
    \toprule
    Compression Rate & 0.0125\% & 0.1\% & 1\% & 10\% \\
    \midrule
    Accuracy & 84.73 $\pm$ 0.76 & 88.60 $\pm$ 0.87 & 89.67 $\pm$ 0.23 & 89.73 $\pm$ 0.31 \\
    \bottomrule
    \end{tabular}
    \caption{Accuracy Experiment 2 (Basic Protocol) for FSL with Mega-Elements}
    \label{tab:mega-element}
\end{table*}

\subsection{Real-world Tasks with FSL}
\label{sec:real-world-tasks}

We validated the effectiveness of top-$k$ (with dynamic submodels to be aggregated for each communication round)
on different datasets \cite{aji2017sparse}.
We consider two widely used federated image datasets MNIST\cite{lecun1998gradient} and CIFAR10\cite{krizhevsky2009learning} and an NLP text materials dataset TREC\footnote{Test REtrieval Conference (TREC) dataset: \url{https://trec.nist.gov/data.html}}. 
For the MNIST and CIFAR10 dataset, we follow the IID data partitioning strategy described in \cite{mcmahan2017communication}, where split the two datasets evenly into 100 clients after being randomly shuffled respectively. We adopt the CNN model architecture used in \cite{mcmahan2017communication} for MNIST and CIFAR10. 
As for TREC dataset, following \cite{zhu2020empirical}, the dataset is distributed to 4 clients evenly. 
The statistics of the dataset is shown in Table  \ref{tab:trec-statistics}, and we adopt the same TextCNN \cite{Kim2014ConvolutionalNN} model, which is a convolutional neural network with additional embedding layer for text classification.

\begin{table}[t!]
    \centering
    \begin{tabular}{@{}lcccccccccc@{}}
    \toprule
    Type   & \#Client (s) & \#Words & \#Samples \\
    \midrule
    Test (Full) & / & 753 & 500 \\
    Train (Full) & 4 & 8256 & 5452 \\
    Train (Per Client) & 1 & 3365 & 1363 \\
    \bottomrule
    \end{tabular}
    \caption{Statistics of TREC dataset.}
    \label{tab:trec-statistics}
\end{table}

For MNIST/CIFAR10, the models are trained with an SGD optimizer and we set the learning rate as 0.01 and 0.1 with a learning rate decay of 0.99 per 10 rounds respectively, weight decay as 5e-4, batch size as 50, client participation per round as 10\%, local iteration as 1 and global communication rounds as 6000 and 5000 respectively. For TREC, the models are trained with Adam optimizer and we set the learning rate as 0.001, batch size as 64, client participation per round as 100\% (full participation), local iterations as 2 and global communication rounds as 500.

All accuracy results are reported with mean and standard deviation over three random runs, which are reported in Table \ref{tab:naive-topk}.
Experimental results show that on each task, when the compression rate exceeds a certain threshold, as the compression rate increases, the final convergence accuracy of the model will experience a slight drop, which is tolerable considering that the communication cost can be reduced a lot with such top-$k$ submodel selection strategy. In addition, we observe that when the compression ratio reaches 50 times, the performances of the models on the MNIST, CIFAR10, and TREC datasets are only reduced by 0.11\%, 0.78\% and 0.73\%, respectively. This demonstrates that the top-$k$ strategy is effective and practical for the federated learning.

\subsection{FSL with Mega-Elements}

As demonstrated in the previous section, top-$k$ selection strategy using the basic protocol results in similar performance (in terms of accuracy) as the full model training. In this section, we conduct an additional experiment on the TREC dataset to validate our proposed mega-element optimisation and the experimental settings are consistent with those described in section \ref{sec:real-world-tasks}. 

We perform top-$k$ mega-element selection operation on the parameters of the word embedding layer. In detail, parameters in each row of the word embedding layer are grouped (sum all the elements' absolute values) into a number to represent the importance of the word, and then select the top-$k$ mega-elements among all words.

The experimental results are shown in Table \ref{tab:mega-element}. Note that the compression ratio here is calculated in comparison with the parameters of the entire embedding layer instead of the entire neural network.
From the results, we can conclude that the top-$k$ mega-elements selection strategy is robust against a wide range of compression rate. 
%

\subsection{Comparing with Existing Scheme}
Now, we would like to compare our method with the most recent FSL aggregation method with \textit{differential privacy} to protect the submodel information~\cite{niu2020billion}.
For a fair comparison, we adopt the same model, i.e., Deep Interest Network (DIN)~\cite{zhou2018deep} for a industrial-scale recommendation task presented in ~\cite{niu2020billion}.
DIN consists of embedding layers, attention layers, and fully connected layers. 
More specifically, their model contains a total of 3,617,023 parameters, where the embedding layers account for 3,552,696 of the parameters (i.e., 98.22\% of the whole model) and the rest accounts for only 64,327. 
From the perspective of the client, each client interacts with an average of only 301 goods IDs and 117 category IDs, which means that only 7,542 embedding parameters are updated per client.
In addition to the parameters of the remaining components, the desired size of the sub-model on the client is 71,869.
While using 128-bit float-point representation, the submodel at the client side is 1.09MB.
And with the PSU protocol as the additional cost, the communication overhead per client per round is at least 1.76MB.
With the same model, each client in our basic SSA protocol uploads 1.4MB embedding layer with 0.98MB other components.
As for total round time, each client finishes one round within 3s (except for local training), each server finishes aggregation within 1 min.

\section{Conclusion}
In this paper, we proposed basic PSR and SSA protocol for secure federated submodel learning in the \textit{semi-honest and malicious} adversary models.
We also proposed several optimisations techniques to improve protocol efficiency and its practicality for real-world tasks.
Also, we use experiments to demonstrate the practical efficiency of our protocol and show that our protocol outperforms existing solutions and the full-blown secure aggregation protocols.

\bibliographystyle{plain}
\bibliography{main}


\end{document}